\documentclass{article}

\usepackage[preprint]{neurips_2026}

\usepackage[utf8]{inputenc}
\usepackage[T1]{fontenc}
\usepackage{url}
\usepackage{booktabs}
\usepackage{xspace}
\usepackage{amsfonts}
\usepackage{amsthm}
\usepackage{amsmath}
\usepackage{amssymb}
\usepackage{nicefrac}
\usepackage{microtype}
\usepackage{xcolor}
\usepackage{graphicx}
\usepackage{subcaption}
\usepackage{tabularx}
\usepackage[export]{adjustbox}

\usepackage{hyperref}
\usepackage[capitalize]{cleveref}

\newcommand{\R}{\mathbb{R}}

\newcommand{\Row}[1]{\mathsf{Row}_{#1}}
\newcommand{\Col}[1]{\mathsf{Col}_{#1}}
\newcommand{\Bx}[1]{\mathsf{Box}_{#1}}
\newcommand{\cands}{\mathsf{cand}}

\newcommand{\dmodel}{d_{\mathrm{model}}}
\newcommand{\dmlp}{d_{\mathrm{mlp}}}

\newcommand{\pvec}[2]{w_{#1,#2}}

\newcommand{\cluesend}{\texttt{[clues\_end]}\xspace}
\newcommand{\push}{\texttt{[push]}\xspace}
\newcommand{\pop}{\texttt{[pop]}\xspace}
\newcommand{\success}{\texttt{[success]}\xspace}

\theoremstyle{definition}

\theoremstyle{plain}

\theoremstyle{remark}

\title{Transformers Linearly Represent\\ Highly Structured World Models}

\author{  Roman Kniazev \\
  LaBRI, CNRS\\
  University of Bordeaux\\
  France \\
  \texttt{roman@knzv.me} \\
  \And
  Nathana\"el Fijalkow \\
  LaBRI, CNRS\\
  University of Bordeaux\\
  France \\
  \texttt{nathanael.fijalkow@gmail.com}}

\begin{document}

\maketitle

\begin{abstract}
Do transformers, when trained on sequential reasoning traces, build internal models of
the underlying task?
And if so, does the structure of those internal representations mirror the structure of
the domain?

We train an 8-layer transformer on Sudoku solving traces and perform a mechanistic
analysis of its internal computation.
We establish two results.
First, the model builds a \emph{substructure world model}: it does not represent the
board state cell by cell, as a human analyst would expect, but organizes information
around the rows, columns, and boxes that Sudoku's constraints act on.
Second, we identify a \emph{naked-single circuit}: a small set of dedicated neurons in
the final MLP layer, each individually detecting when exactly one digit remains
possible for a specific cell, and reliably promoting that digit.

These findings show that the geometry of an emergent world model is shaped by the
constraint algebra of the domain, not its surface presentation, and that the resulting
decision circuit is sparse, monosemantic, and fully interpretable.
More broadly, they demonstrate that mechanistic interpretability tools can recover
an end-to-end algorithmic account of how a transformer solves a combinatorial
reasoning task.
\end{abstract}

\section{Introduction}
\label{sec:intro}

\paragraph{Context.}
Do transformers, when trained on sequential reasoning traces, build internal
\emph{world models} of the underlying environment?
\citet{li2023emergent} and \citet{nanda2023emergent} showed that the answer is yes for Othello: a GPT-2-style model
trained only on game transcripts spontaneously develops a representation of board
occupancy (black / white / empty per cell), despite receiving no direct supervision on
board state.
This finding sits within the programme of mechanistic interpretability 
\citet{elhage2021mathematical, elhage2022toy}, which aims to reverse-engineer the 
algorithms learned by neural networks. 
A central working hypothesis in this programme is the linear representation hypothesis 
\citet{park2024linear, mikolov2013linguistic}: that high-level concepts are encoded as linear directions in activation space, and can therefore be read out by simple linear probes \citet{alain2016understanding, belinkov2022probing}. Understanding when and how world models emerge under this hypothesis, and what geometry they take, is important for building reliable, inspectable AI systems.

\paragraph{Background: linear world models and how to find them.}
\citet{nanda2023emergent} built on \citet{li2023emergent}'s result by establishing that
the Othello world model is \emph{linear}: the occupancy of each cell is recoverable by
a simple logistic regression \emph{probe} on the residual stream, without any nonlinear
transformation.
They further showed, using \emph{activation patching}, that these linear directions are
causally used by the model to predict the next move, not merely a side effect of
training.
The same paradigm has since revealed world models in chess-playing
transformers~\citep{karvonen2024emergent} and maze-solving
models~\citep{ivanitskiy2024linearly}.
In all of these cases, the unit of representation matches the most natural human
decomposition of the state space: one feature per cell.

\paragraph{Our question: does the geometry reflect the task's constraint algebra?}
We ask whether this phenomenon extends to a qualitatively different class of task:
combinatorial \emph{constraint-satisfaction}.
Our domain is \textbf{Sudoku}: a one-player puzzle on a $9 \times 9$ grid whose
fundamental difficulty is constraint propagation across 27 overlapping
\emph{substructures} (9 rows, 9 columns, 9 boxes), each requiring every digit $1$--$9$
to appear exactly once.
Unlike Othello, where the state is naturally decomposed by cell, Sudoku has an
explicit algebraic structure: the validity of any placement is determined by the
digit's presence in three substructures, not in any single cell.
This raises a sharp question: does the model represent the board state \emph{per cell}
(as in Othello) or \emph{per substructure} (as the constraints demand)?

Following \citet{giannoulis2026teaching} and \citet{shah2024causal}, we train a causal transformer on Norvig-style
solving traces with backtracking~\citep{norvig2006sudoku}, achieving 98.4\% per-cell and
97.5\% per-grid accuracy on a held-out test set of 150{,}000 puzzles.
We then apply \citeauthor{nanda2023emergent}'s probe-and-patch methodology to the
residual stream.

\paragraph{Findings and contributions.}
Our contribution is new findings about the geometry of emergent world models 
and the circuits that read them, obtained by applying the probe-and-patch 
methodology of \citet{nanda2023emergent} to a transformer trained on Sudoku 
following \citet{giannoulis2026teaching}.

\begin{enumerate}
  \item \textbf{Substructure world model} (\cref{sec:latent}).
Prior world-model work has consistently found representations whose geometry 
matches the surface decomposition of the task: one feature per cell. 
We show this is not a universal property: when the task has an explicit 
constraint algebra, the model's representation reflects that algebra instead. 
Linear probes for ``is digit $d$ present in substructure $S$?'' achieve 
perfect exact match accuracy across all 243 (substructure, digit) pairs 
in mid-layers, while per-cell probes plateau at 80\% accuracy. 
The representation forms by mid-layers and tracks the current board state 
throughout generation, remaining geometrically stable across the entire 
solving trace.
Causal patching confirms these directions are actively~used.

  \item \textbf{Substructure-aligned attention} (\cref{sec:mid-layers}).
We identify that mid-layer attention heads organize along Sudoku's substructures: 
heads concentrate their attention on individual rows, columns, or boxes 
(or coherent groups of three), 
and their OV circuits apply a symmetric mechanism:
digits already present in the attended substructure have their logits suppressed, 
while absent digits are promoted. 
This shows that the transformer represents Sudoku's rules through localized 
constraint checkers operating directly on the substructure-organized residual stream.

  \item \textbf{Naked-single circuit} (\cref{sec:decision}).
We identify a small set of dedicated neurons in the final MLP layer that detect
when exactly one digit remains possible for a specific cell (a \emph{naked single}).
Each neuron is specific and monosemantic. Moreover, the residual stream before the final MLP already strongly promotes the correct digit, so mid-layers encode the answer and the final MLP amplifies it.
The circuit is sparse and provides a fully mechanistic account of
how the model handles the most common forced placement.
\end{enumerate}

Code is available at \href{https://github.com/kameronton/sudoku-residual}{the link},
and the dataset and a model checkpoint at \href{https://huggingface.co/datasets/residual-sudoku-dataset/residual-sudoku-dataset}{Hugging Face}.

\section{Preliminaries}
\label{sec:prelim}

\paragraph{Sudoku.}
A Sudoku puzzle is defined on a $9 \times 9$ grid of 81 cells.
Each cell $(r,c)$, for $r,c \in \{1,\ldots,9\}$, is to be filled with a digit
$d \in \{1,\ldots,9\}$.
The grid is partitioned into 27 \emph{substructures}: 9 rows, 9 columns, and 9 \emph{boxes}.
Row $r$ is $\Row{r} = \{(r,c) : c \in \{1,\ldots,9\}\}$;
column $c$ is $\Col{c} = \{(r,c) : r \in \{1,\ldots,9\}\}$;
box $\Bx{k}$ for $k = 3(i-1)+j$ ($i,j \in \{1,2,3\}$) is the $3\times 3$ block
$\{(3i-2,3j-2),\ldots,(3i,3j)\}$.
Every cell $(r,c)$ belongs to exactly three substructures: $\Row{r}$, $\Col{c}$, and one box $\Bx{k}$.
Three horizontally adjacent boxes sharing the same rows form a \emph{band};
three vertically adjacent boxes sharing the same columns form a \emph{stack}.
There are 3 bands and 3 stacks, each covering 27 cells.

A placement of digit $d$ in cell $(r,c)$ is \emph{valid} if $d$ does not already appear
in $\Row{r}$, $\Col{c}$, or $\Bx{k}$.
The \emph{candidate set} $\cands(r,c) = \{1,\ldots,9\} \setminus (\text{digits in }\Row{r}\cup\Col{c}\cup\Bx{k})$
collects all valid digits for an empty cell.
A puzzle is given as an initial partial assignment (the \emph{clues}), and the task is
to complete it uniquely so that every digit appears exactly once per substructure.

\paragraph{Solving techniques.}
Two basic deductive rules eliminate candidates before backtracking search is needed.
Cell $(r,c)$ is a \emph{naked single} if $|\cands(r,c)| = 1$; the unique candidate must
be placed there.
Digit $d$ is a \emph{hidden single} in substructure $S$ if exactly one empty cell
$(r,c)\in S$ has $d\in\cands(r,c)$; that cell must receive $d$.
When neither rule applies, a \emph{backtracking search} is used: the solver selects the
cell with the fewest candidates, guesses a digit, and rolls back if a contradiction is
reached.

\paragraph{Solving traces and model vocabulary.}
Following~\citet{giannoulis2026teaching} and~\citet{shah2024causal}, we generate solving
traces using a Norvig-style solver~\citep{norvig2006sudoku} with several randomizations
(order of constrained placements, backtracking cell selection, digit order), producing
diverse traces per puzzle.
The vocabulary consists of \textbf{729 placement tokens}
$\mathtt{[R}r\mathtt{C}c\mathtt{=}d\mathtt{]}$ for each cell and digit, plus special
tokens \cluesend (end of clues), \push (backtracking branch start), \pop (failed
branch), \success (puzzle solved), and \texttt{[pad]}.
A typical trace takes the form:
\begin{multline*}
  \underbrace{
    \mathtt{[R1C1=5]\ \cdots\ [R2C4=3]}
  }_{\text{clues}}
  \ \cluesend\
  \underbrace{
    \mathtt{[R3C5=9]\ [R6C8=4]}
  }_{\text{deductions}}
  \ \push\ \cdots \\
  \cdots \mathtt{[R8C8=1]}\ \pop\
  \push\ \mathtt{[R8C8=2]}\ \cdots\ \success.
\end{multline*}
The \cluesend token serves as the primary anchor for our probing experiments:
it follows all clue tokens and precedes any model-generated placement, providing a
clean position at which to measure the model's representation of the initial board state.

\paragraph{Model architecture and training.}
We train a standard causal decoder-only transformer following~\citet{giannoulis2026teaching}:
$L = 8$ layers, $H = 8$ attention heads, pre-layer normalization, $\dmodel = 576$,
$\dmlp = 3{,}456$ (expansion ratio $6\times$).
The model is trained on 2.7M puzzles from the sudoku-3m dataset~\citep{sudoku3m}, with
traces trimmed to 250 tokens (only 0.9\% are longer), for 6 epochs
on a Google TPU v6e, achieving \textbf{98.4\% per-cell} and
\textbf{97.5\% per-grid} accuracy on a held-out test set of 150K puzzles.

\paragraph{Linear probing.}
A \emph{linear probe}~\citep{alain2016understanding} for a binary concept $f$ is a
logistic regression classifier trained on residual stream activations $x \in
\R^{\dmodel}$: 
\[
P(f = 1 \mid x) = \sigma(\pvec{f}{} \cdot x + b_f).
\]
The probe direction $\pvec{f}{} \in \R^{\dmodel}$ recovers a linearly accessible
encoding of $f$ in the model's internal representation.
A near-perfect probe (AUC $\approx 1$) certifies that the concept is linearly encoded;
an imperfect one does not rule out encoding at a different level of abstraction.
Unless otherwise stated, probes are trained on 5{,}120 puzzles from the test set and
evaluated on 1{,}280, probing activations at the \cluesend position after each layer.

\paragraph{Causal intervention.}
Linear probes identify \emph{where} information is encoded but not whether the model
\emph{uses} it.
Activation patching~\citep{geiger2021causal} answers this causal question: we subtract
a scaled probe direction from the residual stream at a given layer and measure the
change in output logits.
If the intervention disrupts the correct prediction, the direction is causally
involved in the computation, not merely a side effect of training.

\paragraph{Direct logit attribution and the logit lens.}
To track how predictions build up across layers, we use \emph{direct logit
attribution}~\citep{elhage2021mathematical} and the \emph{logit
lens}~\citep{nostalgebraist2020logit}.
The logit lens projects the residual stream after each layer through the final
unembedding matrix, reading off the model's ``current best guess'' at every
intermediate step.
Direct logit attribution decomposes the final logit difference into additive
contributions from each attention head and MLP layer, identifying which components
are responsible for promoting the correct token.

\paragraph{Attention head decomposition.}
Each attention head can be factored into two independent circuits~\citep{elhage2021mathematical}.
The \textbf{QK circuit} (query--key matrices) determines \emph{which} positions the
head attends to.
The \textbf{OV circuit} (output--value matrices) determines \emph{what} information is
read from those positions and written into the residual stream.
Analyzing them separately lets us ask: \emph{what does the head look at?} (QK) and
\emph{what does it do with what it sees?} (OV).

\section{Understanding the Latent Space}
\label{sec:latent}

\begin{figure}[t]
     \centering
     
     \begin{subfigure}[b]{0.48\textwidth}
         \centering
         \includegraphics[width=\textwidth]{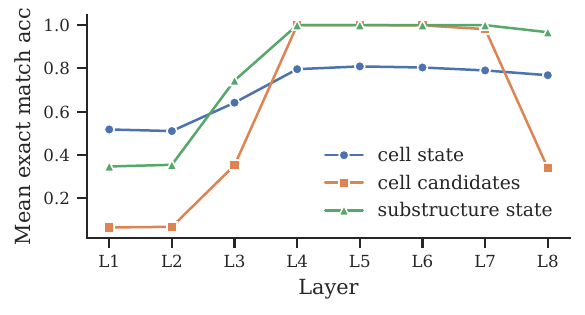}
     \end{subfigure}
     \hfill
     \begin{subfigure}[b]{0.48\textwidth}
         \centering
         \includegraphics[width=\textwidth]{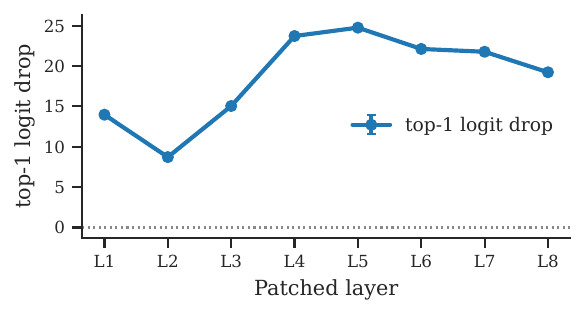}
     \end{subfigure}
     
     \caption{
    \textbf{(Left)} Mean exact match accuracy of three families of linear probes trained at \cluesend token over different layers.
    \emph{Blue}: 81 multi-class probes predicting the digit in a cell (top-1 accuracy);
    \emph{Orange}: 729 binary probes predicting if a digit is a valid candidate in a cell (exact match across cell);
    \emph{Green}: 243 binary probes predicting if a digit is present in a substructure (exact match over a substructure);
    \textbf{(Right)} Logit score drop of the top-1 target token after activation patching of substructure probe directions across layers. 
    }
    \label{fig:probes}
\end{figure}

\subsection{Cell-Level Representation}

Prior studies into world representation in sequence models trained 
on grid-based environments, such as OthelloGPT (\cite{nanda2023emergent,li2023emergent}),
have based the analysis on cell-based features, as it is the smallest feature scale 
of the environment.
Given that Sudoku is presented visually and structurally as a $9\times 9$ grid, 
the natural baseline hypothesis is that the network learns a similar spatial decomposition,
representing the state of the board via independent, cell-local features.

\paragraph{Are the contents of cells linearly represented?}
To test this spatial hypothesis, we apply linear probing at the cell level. 
We train 81 independent multi-class linear classifiers (one for each cell $(r,c)$ 
of the grid) on the residual stream activations at the \cluesend token. 
Each probe acts as a 9-class classifier trained to predict the digit occupying its respective cell.
The probes are trained independently on the residual stream after each of eight layers.

As the main performance measure, we use top-1 accuracy,
which is shown in \cref{fig:probes} (left).
The probes perform significantly better than random guessing, 
but they fail to achieve perfect linear separability. 
The accuracy peaks after layer 4 at $0.8$ and stays stable through layers 5, 6 and 7.
We additionally measure the MSE between the predicted probability vectors and the one-hot targets,
(\cref{app:mse-cell-probes}).
The probes do better than chance but never reach perfect accuracy, suggesting that cell state is not cleanly linearly encoded in the residual stream.

\paragraph{Are the candidates linearly represented in each cell?}
In OthelloGPT, the linear representation of the board only emerged once the probe target was reframed from absolute (black/white) to relative (mine/theirs).
Following this, we hypothesize that the network primarily works with the candidates instead of filled digits. 
In this experiment, we maintain the cell as our spatial unit but shift the predicted concept to candidate sets.

We train 729 binary linear probes to predict the valid candidates of each cell,
that is, each probe answers ``If cell $(r, c)$ is empty, is digit $d$ a valid candidate in it?''. 
In stark contrast to the state probes, these candidate-set probes achieve 
$1.0$ exact match accuracy through layers 4 to 6 when predicting 
the full 9-digit candidate set for any given cell, as can be seen in \cref{fig:probes} (left).

To check if the 729 cell-candidate probes encode 729 independent features, we extract the normal vectors of the 81 probes for a fixed digit and compute their pairwise cosine similarity at layer 6.
We find that the vectors are not independent.
Two cells that share a substructure have probes aligned at $\approx\!0.33$
on average; two cells that share two substructures (e.g.\ a row and a box, or a column and a box) align at $\approx\!0.65$. Vectors for cells with no shared substructure are near-orthogonal. Other mid-layers show the same pattern with greater spread (\cref{app:probe-cosine-sim}).
This structural alignment strongly implies that the model does not maintain 81 isolated candidate sets, 
but instead the cell-level candidate vectors are linear combinations of substructure-level features.

\subsection{Substructure-Level Representation}

\paragraph{Are the candidates linearly represented in each substructure?}
The structural colinearity of the cell-candidate vectors suggests that the network 
factorizes the board state according to the puzzle rules:
whether a digit is a valid candidate in a cell is dictated not by the cell itself, 
but entirely by absence of that digit in the row, column, or box the cell belongs to.

We test whether the representation uses this algebraic structure by shifting 
the unit of analysis from the cell to the logical constraint group. 
We train 243 binary linear probes that predict a global constraint: 
``Is digit $d$ present in substructure $S$?'' (covering all 9 rows, 9 columns, and 9 boxes for each of the 9 digits).
As shown in \cref{fig:probes} (left), these substructure probes achieve perfect exact match accuracy in mid-layers,
meaning that the probe correctly predicts the precise status of a digit across all 27 substructures simultaneously.
The score doesn't degrade in the last layer, contrary to cell-level candidate probes.

There is a useful asymmetry here. If a digit is absent from the row, column, or box of a cell, it cannot be a candidate there, so substructure-absence linearly determines cell-candidacy. The converse does not hold: a digit being present in all three substructures does not mean it is the digit placed in the cell, since the cell may simply be empty. This explains the gap between our two probe families: candidates are a linear function of substructure state, while filled-cell state is not.

\subsection{Causal Ablation}

\paragraph{Is the substructure-level representation actually used for prediction?}
We have shown that substructure directions can be decoded from the residual stream. 
Are they actually used by the model when it predicts the next placement? 
We test this by overwriting those directions to encode that the digit is already present in the row, column, and box of a target cell and check whether the model's prediction for that placement collapses.

Concretely, we collect 300 grid states (G1) at \cluesend,  
together with the consecutive next step (G2) in which a cell $(r,c)$ is filled with digit $d$.
We then patch the residual stream of (G1) at a given layer by transplanting 
the $(\Row{r}, d)$, $(\Col{c}, d)$, and $(\Bx{k}, d)$ probe direction components from (G2) into (G1):
\begin{equation}
    x'_{G1} = x_{G1} + \sum_{S \in \{\Row{r},\, \Col{c},\, \Bx{k}\}} 
    \bigl(\hat{w}_{S,d}^\top x_{G2} - \hat{w}_{S,d}^\top x_{G1}\bigr)\, \hat{w}_{S,d}
\end{equation}
where $\hat{w}_{S,d}$ is the unit-normalized probe direction for substructure $S$ and digit $d$.
We then measure the drop in the logit of the target token $\mathtt{[R}r\mathtt{C}c\mathtt{=}d\mathtt{]}$, 
which under the clean forward pass is the model's top prediction (mean clean logit 16.82).

As shown in \cref{fig:probes} (right), the target logit drops across all layers, peaking at layer 4 with a mean drop of approximately 25 logits, and remaining high through layers 5--8. 
The intervention is causally effective: the patched model's top-1 prediction differs from the unpatched prediction in 99\% of cases, confirming the directions are used. 
The patched top-1 is itself a valid placement on the original board in 73--76\% of cases at L3--L4,
suggesting that the patched state remains coherent. Full table can be found in \cref{app:structure-present-patch}.

\subsection{Transfer Across Solving Steps and Layers}

\begin{figure}[t]
     \centering
     
     \begin{subfigure}[b]{0.48\textwidth}
         \centering
         \includegraphics[width=\textwidth]{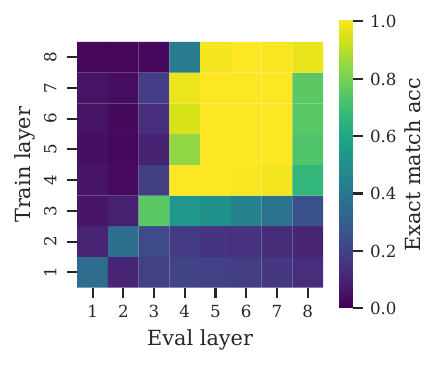}
     \end{subfigure}
     \hfill
     \begin{subfigure}[b]{0.48\textwidth}
         \centering
         \includegraphics[width=\textwidth]{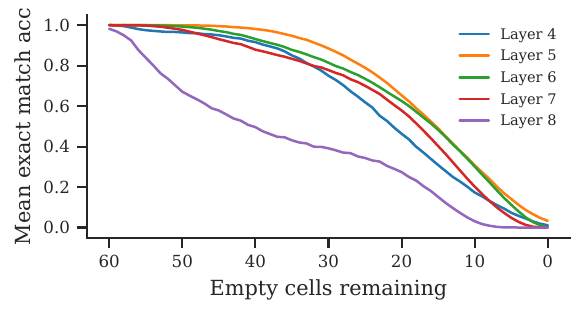}
     \end{subfigure}
     \caption{
        \textbf{(Left)} Cross-layer transfer of substructure-state probes: entry $(i,j)$
        is the mean exact match accuracy of probes trained on layer $i$ activations and
        evaluated on layer $j$ activations without retraining;
        \textbf{(Right)} Cross-position transfer of substructure-state probes trained on different layers: 
        mean exact match accuracy of probes trained at \cluesend and evaluated on later
        positions without retraining.
        }
     \label{fig:transfer}
\end{figure}

\paragraph{Is the substructure-level representation consistent across layers?}
To identify at which layer the substructure representation stabilizes, we perform a 
cross-layer probe transfer experiment: for each of the 243 (substructure, digit) probes, 
we train on activations at a fixed source layer and evaluate on every other layer without 
retraining. As shown in \cref{fig:transfer} (left), the transfer matrix reveals a sharp 
phase transition at layer 4: probes trained on layers 1--3 fail to transfer to any later 
layer, while probes trained on layers 4--7 achieve near-perfect exact match accuracy 
across the entire mid-layer span.
It means that the (substructure, digit) world model is absent in early 
layers, appears at layer 4, and remains stable through layer 7.

\paragraph{Is the substructure-level representation persistent across solving steps?}
To test if the substructure representation persists dynamically, we apply the frozen 
probes (trained at the \cluesend token) to subsequent positions in the generating trace. 
As shown in \cref{fig:transfer} (right), the exact match accuracy of these probes steadily 
degrades toward zero as the puzzle is solved. 
However, this drop is an artefact of the probes' decision threshold, not the underlying representation. As the grid fills, almost every digit becomes present in almost every substructure, so the optimal threshold shifts, but the probes' biases were fixed at training time on a much sparser distribution.
To isolate the underlying feature directions from this shifting decision threshold, 
we evaluate the ROC-AUC of the exact same probes. 
As detailed in \cref{app:transfer-steps}, the AUC stays at $1.0$ 
across the entire trace for the mid-layers, confirming that the model maintains 
a structurally invariant world model throughout generation.

\section{Mid-Layer Routing and Constraint Elimination}
\label{sec:mid-layers}

In \Cref{sec:latent} we showed that the model's residual stream is organized by Sudoku's substructures. 
In this section, we ask which components write that organisation into the stream. 
We analyze the behavior of mid-layer attention heads across a dataset of 6,400 puzzles by extracting the activations at the \cluesend token and evaluating both the attention distributions ($QK$ circuit) and the direct output to the residual stream ($OV$ circuit).

\subsection{Attention Patterns}
\paragraph{What information are attention heads focusing on?}
We project the average attention weights from the \cluesend token onto the $9\times9$ grid. 
Numerous mid-layer heads heavily concentrate their attention mass on specific substructures:
individual boxes, horizontal bands, or vertical stacks (an example can be seen in \cref{fig:mid-layers}; 
full patterns in \cref{add:full-attention-patterns}). 
The mapping between heads and substructures is not strictly one-to-one: 
some heads attend broadly across a band with stronger concentration on one of its boxes, 
and some substructures are jointly covered by several heads. 
Because these heads sit in the middle of the network, ``attending to position $(r,c)$'' means reading the residual stream that has accumulated at that position over the previous layers, not the raw token embedding for the cell.

\subsection{Heads Direct Logit Attribution}
\paragraph{What role does each attention head play in the decision?}
We hypothesize that this specialized attention allows specific heads to act as 
dedicated constraint managers for their respective substructures. 
To verify this, we evaluate the isolated output of these heads using 
Direct Logit Attribution (DLA): we project each head's contribution to the residual stream through the unembedding matrix to read off its effect on the logits of placement tokens and to see how they update the output probabilities.

\begin{figure}[t] 
     \centering
     
    \includegraphics[width=\linewidth]{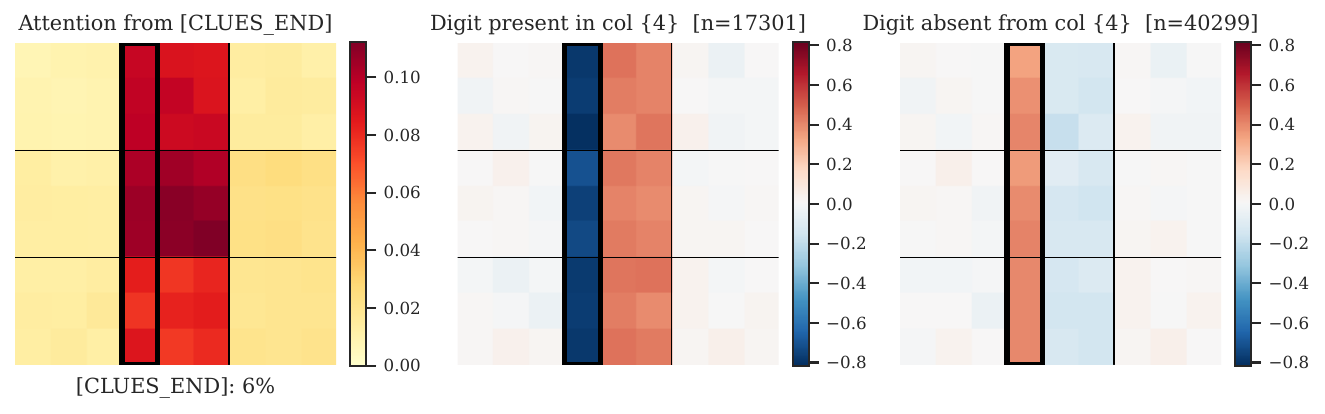}

     \caption{Substructure constraint elimination in mid-layer attention heads. 
     Attention map (left) shows that the head is specialized for routing information 
     from tokens representing placements in middle columns. 
     Direct Logit Attribution (middle and right) shows that the head 
     suppresses candidate logits for digits that present in its target substructure and promotes digits that are absent.
    }
     \label{fig:mid-layers}
\end{figure}

We illustrate it with a stack-attending head (\cref{fig:mid-layers}). 
For each digit $d$ and each column within the stack, we split examples by whether $d$ already appears in that column, and average the head's logit contribution across digits and examples.
We find that the head implements a symmetric mechanism. When $d$ is present in the column, the head suppresses the logit of $d$ across all cells of that column and slightly promotes $d$ in the other columns of the stack. When $d$ is absent, the head promotes $d$
across the column. The same pattern holds for heads specialized on rows, boxes, bands, and stacks (\cref{app:more-heads-dla}).

\subsection{Causal Ablation}

If a head suppresses digits already present in its target 
substructure, then removing it should make the invalid digits more plausible 
to the model. 
We test this with mean ablation on three heads.

For each of the three heads
we replace its output at the \cluesend position with its mean over a held-out set of 1,280 puzzles.
We then measure the change in logit of illegal placements $\mathtt{[R}r\mathtt{C}c\mathtt{=}d\mathtt{]}$
where $(r, c)$ is in the head's target substructure and digit $d$ already appears there.
We additionally measure the effect on a substructure of the same type that the head does not specialize in.

As shown in \cref{tab:head-substructure-ablation}, ablating each head raises logits of illegal 
placements in its target substructure by 0.94--1.69, 
and lowers them slightly in the control. 
Thus, the suppression mechanism is implemented specifically and locally, with each head enforcing constraints in its own substructure,
and removing it does not disrupt constraints of other substructures.

\begin{table}[t]
\centering
\begin{tabular}{llrr}
\hline
Head & Region & Target $\Delta$logit & Control $\Delta$logit \\
\hline
L4H6 & cols 4--6 & $1.426 \pm 0.004$ & $-0.170 \pm 0.002$ \\
L5H8 & rows 7--9 & $1.690 \pm 0.004$ & $-0.216 \pm 0.002$ \\
L6H3 & box 5     & $0.937 \pm 0.005$ & $-0.108 \pm 0.003$ \\
\hline
\end{tabular}
\caption{
Mean-ablation suppression gaps for specialized attention heads.
Values are mean changes in illegal-placement logits,
$\mathbb{E}[\ell_{\mathrm{ablated}} - \ell_{\mathrm{clean}}]$,
reported as mean $\pm$ standard error. Positive $\Delta$target indicate that
ablating the head removes suppression in the substructure.
}
\label{tab:head-substructure-ablation}
\end{table}

\section{Understanding the Decision Layer}
\label{sec:decision}

In \Cref{sec:mid-layers}, we demonstrated that the mid-layers accumulate constraint 
representations in the residual stream: valid candidate placements are promoted, 
while invalid ones are suppressed.

In this section, we show that the final decision is implemented in the MLP block of 
the last layer through a two-stage mechanism:
mid-layers accumulate the constraints, and a sparse set of neurons in the final MLP
gates the commitment.
We examine this mechanism on the most common placement type: naked singles (NS).

\subsection{Logit Scores Before the Last MLP Block} 
\begin{figure}[t]
     \centering
     
     \begin{subfigure}[b]{0.48\textwidth}
         \centering
         \includegraphics[width=\textwidth]{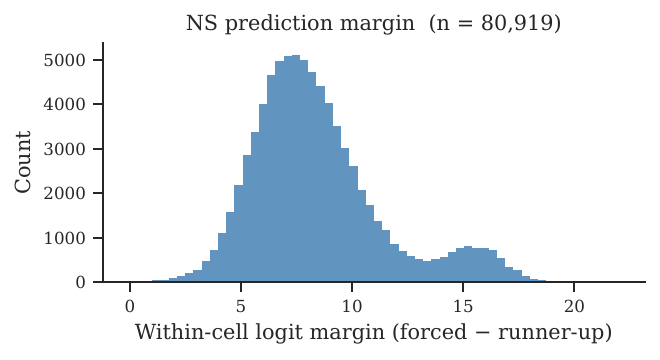}
     \end{subfigure}
     \hfill 
     \begin{subfigure}[b]{0.48\textwidth}
         \centering
         \includegraphics[width=\textwidth]{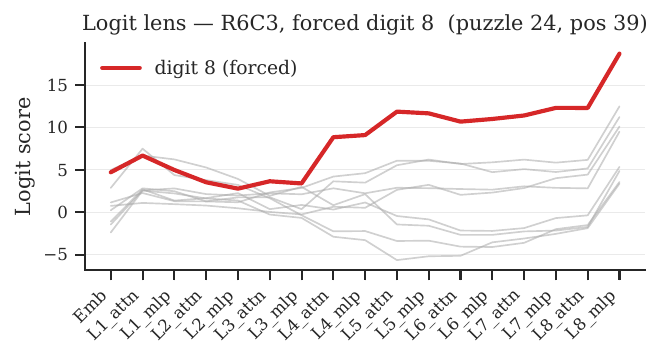}
     \end{subfigure}
     
     \caption{
    \textbf{(Left)} The distribution of the within-cell logit margin (correct digit score - max incorrect digit score) across 80,919 states with unique NS placement.
    \textbf{(Right)} A representative logit lens trace of tokens of a NS cell. The correct digit candidate (red) separates from other digits of the cell (gray) across the layers, with a general promotion of the logits for all digits of the cell in the last MLP block.}
     \label{fig:ns}
\end{figure}

To establish the state of the network before MLP's intervention, 
we apply a logit lens to the residual stream immediately following the L8 attention 
block for all states containing a unique NS.
From all 487,725 states, we extract 1,073,757 target NS placements (since one grid may contain
several NS simultaneously) and then rank the placement tokens for the target cell 
based on their raw logit scores.
We observe that the correct digit is ranked first in 98.77\% of cases.
The distribution of margin scores (the logit score of the correct digit for the cell 
minus the logit score of the runner-up) for the 80,919 states where an NS is unique 
can be seen in \cref{fig:ns} (left).
It shows that before the residual stream reaches the final MLP block, it already 
contains the correct guess by a wide margin.

\subsection{Final Layer Promotion}

To determine how the network decides NS placements, we first manually inspect 
the logit trajectories. 
As shown in \Cref{fig:ns} (right), we observe that the final MLP block applies 
a massive, simultaneous logit promotion to all digit tokens associated with 
a cell once it reaches a unique candidate state.

In order to isolate the mechanism driving this boost, we scan the final MLP for cell-specific neurons. 
For each (cell, neuron) pair, 
we compute the neuron's mean post-activation conditioned on 
the number of valid candidates remaining in that cell,
and define the \emph{activation gap} as
the difference between the neuron's mean activation when the cell has exactly one candidate (i.e. an NS state) 
and its maximum mean activation across other candidate counts.

The distribution of these gaps shows that the vast majority of cell-neuron pairs cluster near zero. 
However, there's a small high-magnitude population separated from this noise floor. 
Setting a cutoff at $3.0$ isolates 91 neurons.
These neurons act as dedicated naked single detectors: they fire exclusively when 
their associated cell is an NS, and remain unactivated otherwise (\cref{fig:ns-neuron}).
The coverage is complete:
71 cell have a single associated neuron and 10 cells have two.

We quantify the effect of these 91 neurons on the final prediction by applying DLA to their output weights directly: rather than measuring per-example contributions as in \Cref{sec:mid-layers}, we project each neuron's output weight vector through the unembedding matrix (folding in the final LayerNorm scale) to read off the logit contribution it makes whenever it fires.
We find that the NS-detector neurons 
output a massive vector that boosts logits of all tokens associated with the cell.
On average, a firing neuron applies a +3.00 logit boost strictly to the placement tokens of its associated cell. 
The variance of this boost across the 9 target digits is small ($\sigma=0.10$), 
which shows that the amplification is essentially digit-uniform. 
Furthermore, the projection onto all other empty cells on the board is negligible ($\mu=-0.02,\sigma=0.03$). 
Additional statistics are listed in \cref{app:ns-dla}.

Together, these results describe the full circuit for NS placements:
first, the mid-layers rank the unique valid candidate first among the 9 placements for the cell;
second, the cell's NS neuron in the final MLP detects the unique-candidate state and uniformly boosts all 9 placement logits for that cell. Since the correct digit was already top-ranked, the uniform boost preserves its lead and the softmax concentrates probability on it.

\subsection{Causal Ablation}

\begin{figure}[t]
    \centering
    \begin{minipage}[b]{0.48\linewidth}
        \centering
        \includegraphics[width=\linewidth]{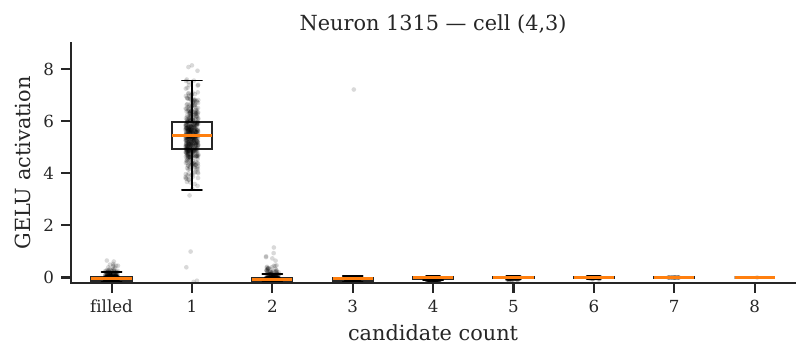}
        \caption{The distribution of activations of an NS neurons.}
        \label{fig:ns-neuron}
    \end{minipage}
\hfill %
    \begin{minipage}[b]{0.5\linewidth}
        \centering
        \scriptsize 
        \begin{tabular}{lcc}
            \hline
            Placement & Logit drop & Probability drop \\
            \hline
            Target NS placement  & $11.408 \pm 0.084$ & $0.585 \pm 0.003$ \\
            Other NS placements  & $-0.655 \pm 0.008$ & -- \\
            \hline
        \end{tabular}
        \captionof{table}{
            Final-MLP naked-single neuron ablation. Values are clean minus ablated scores,
            reported as mean $\pm$ standard error over 1000 evaluation states.
        }
        \label{tab:last-mlp-ns-ablation}
    \end{minipage}

\end{figure}

If an NS neuron is what commits the model to a placement, then ablating it should drop that placement's probability, but leave other naked single placements on the same board untouched.
This sets up a within-puzzle control: 
on a board with two simultaneous NS cells, we ablate the neuron(s) for one and read off the effect on both in a single forward pass.

We collect 1{,}000 such held-out states and mean-ablate the neuron(s) 
associated with the cell the model is about to predict.
As shown in 
\cref{tab:last-mlp-ns-ablation}, ablating these neurons drops the correct 
digit's logit by 11.4 and its softmax probability by 0.59.
In the clean run 
the model splits its probability mass roughly evenly between the two NS 
placements, so a 0.59 drop on one effectively shifts the probability mass to the 
other. The same ablation barely affects the other NS cell (logit increase 0.66).
Thus, the mid-layer signal ranks the correct digit first; the final-layer neuron is what makes the model commit, and it does so for its own cell only.

\section{Conclusion and Future Work}
\label{sec:conclusion}

We trained an 8-layer transformer on Sudoku solving traces and performed a
mechanistic analysis of its internal computation, establishing three results.

The model builds a \emph{substructure world model}: linear probes for
``is digit $d$ present in substructure $S$?'' achieve near-perfect accuracy
across all 243 (substructure, digit) pairs from mid-layers onward,
while per-cell probes plateau at $0.8$.
The geometry of the representation thus mirrors the puzzle's constraint algebra,
not its surface decomposition into cells.
A sharp phase transition at layer~4 separates the layers that build this
representation from those that use it.

Mid-layer attention heads are \emph{substructure-aligned}: they concentrate
their attention on individual rows, columns, or boxes, and their OV circuits
suppress digits already present in the attended substructure while promoting
absent ones, creating a localized constraint-checking mechanism operating directly on
the substructure-organized residual stream.

The final MLP contains a \emph{naked-single circuit}: 91 monosemantic neurons
that detect when exactly one digit remains possible for a given cell.
The residual stream before the final MLP already promotes the correct digit
in 98.77\% of unique naked-single cases, confirming that mid-layers encode
the answer and the final MLP amplifies and commits to it.

\paragraph{Limitations.}
All results concern a single model architecture (8 layers, 8 heads,
$d_\text{model} = 576$) trained on a single puzzle type.
We do not test whether the substructure world model or the naked-single
circuit appear in models of different sizes or in models trained on
non-backtracking traces.
The attention-head analysis in \cref{sec:mid-layers} characterizes the OV
circuits but does not provide a full account of how attention patterns are
selected (QK circuits).
Finally, our causal evidence for the world-model directions relies on a single
patching protocol; alternative interventions could probe the robustness of
these findings.

\paragraph{Future work.}
The most immediate directions are tracking when the substructure geometry and
the circuits emerge during training, either gradually or through a
grokking-like transition~\citep{power2022grokking}, and testing whether the
same organisational principles arise in transformers trained on other
constraint-satisfaction tasks with explicit algebraic structure.

\bibliographystyle{abbrvnat}
\bibliography{references}

\appendix

\section{Details on the Model and Training}
\subsection{Sudoku solving traces}

Training traces are generated by a Norvig-style solver~\citep{norvig2006sudoku} 
modified to introduce randomization at every choice point. At each step, the solver 
first looks for naked singles (cells with exactly one remaining candidate) and 
emits all of them before considering any other rule; when several naked singles are 
available simultaneously, their order is randomised. Once no naked single remains, 
the solver looks for hidden singles (digits with exactly one remaining position 
in a row, column, or box) and emits them in random order. When neither rule 
applies, the solver picks a random cell among those with the smallest candidate set 
and opens a backtracking branch over that cell's candidates, again in random order; 
failed branches are rolled back via \texttt{[pop]} and the next candidate is tried. 
This randomization produces multiple valid traces per puzzle and prevents the model 
from latching onto a fixed cell ordering or digit ordering at training time.
We generate exactly one trace per puzzle and the randomization in the solver ensures the model cannot rely on a fixed cell 
or digit order.

\subsection{Model architecture and training regime}
\label{app:model}

We train a standard causal decoder-only transformer following the architecture 
of~\citet{giannoulis2026teaching}: $L = 8$ layers, $H = 8$ attention heads, 
pre-layer normalisation, model dimension $\dmodel = 576$, MLP width $\dmlp = 3{,}456$ 
(expansion ratio $6\times$), and GELU activations. We do not use dropout.

The training objective is the standard next-token cross-entropy loss, with the loss 
mask applied only to tokens after \cluesend{} so that the model is supervised on its 
own placements and bookkeeping tokens but not on memorising the input clues. This 
differs from~\citet{giannoulis2026teaching} who applies special loss based on the validity of the next possible tokens; 
we  found the standard masked loss sufficient to reach comparable accuracy on this task.

We train on 2.7M puzzles from the sudoku-3m dataset~\citep{sudoku3m}, with traces 
trimmed to 250 tokens (only 0.9\% of puzzles have longer traces), for 6 epochs. Optimisation uses AdamW with learning rate $10^{-3}$, 
weight decay $0.1$, batch size $512$, and a cosine schedule with $5\mathrm{M}$ tokens 
of linear warmup. Training runs on a single Google TPU v6e and reaches 98.4\% 
per-cell and 97.5\% per-grid accuracy on a held-out test set of 150K puzzles.
\newpage
\section{Representation Probing Materials}

\subsection{MSE and AUC Scores of Probes}
\label{app:mse-cell-probes}
\begin{center}     
     \begin{minipage}{0.48\textwidth}
         \centering
         \includegraphics[width=\textwidth]{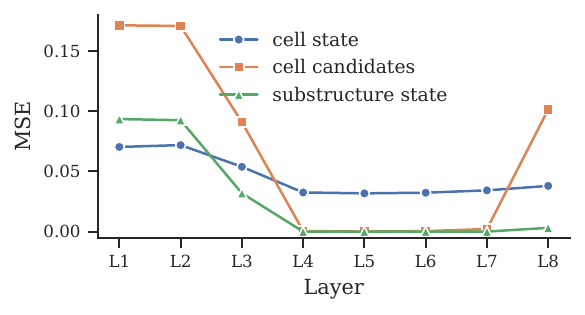}
     \end{minipage}
     \hfill
     \begin{minipage}{0.48\textwidth}
         \centering
         \includegraphics[width=\textwidth]{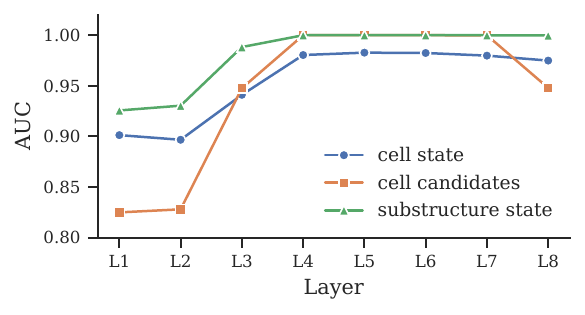}
     \end{minipage}
     
     \captionof{figure}{
\textbf{(Left)} Mean squared error between the probes' predicted probability vectors 
and the one-hot targets, per layer, for the three probe families: 
81 cell-state probes (blue), 729 cell-candidate probes (orange), and 243 substructure-state 
probes (green). Cell-state probes settle around $\mathrm{MSE}\!\approx\!0.03$ in 
mid-layers and never reach zero, consistent with their imperfect top-1 accuracy; 
substructure-state probes drive MSE close to zero from layer 4 onward and stay 
there through the final layer; cell-candidate probes match this in mid-layers but 
deteriorate sharply at L8.
\textbf{(Right)} Mean ROC-AUC of the same three probe families, per layer. 
Substructure-stat probes reach near-perfect AUC by layer 4 and remain at 
$\approx\!1.0$ across all subsequent layers, with cell-candidate AUC dropping at L8. 
Cell-state probes 
plateau slightly below at layers 4--7.}

\end{center}

\subsection{Substructure Alignment of Cell-Candidate Probes}
\label{app:probe-cosine-sim}
\begin{center}
  \centering
  \includegraphics[width=0.6\linewidth]{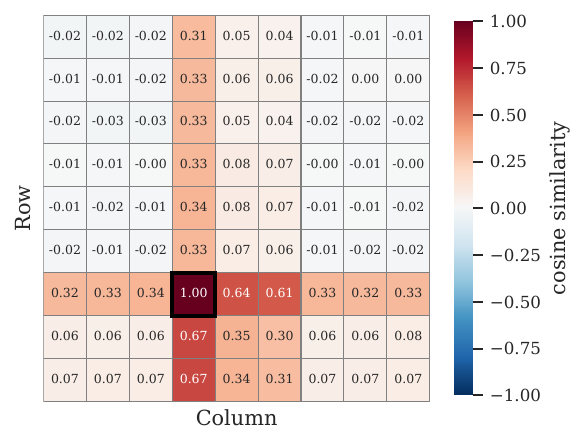}
  \captionof{figure}{Cosine similarities between the probe trained at layer 6 activations to predict if $3$ is a valid candidate
  in cell $(7,4)$ and probes trained on other cells for digit $3$.}
\end{center}

\begin{table}[ht]
\centering
\tiny 
\setlength{\tabcolsep}{4pt} 
\begin{tabularx}{\textwidth}{l *{8}{X}}
\toprule
\textbf{Category} & \textbf{L3} & \textbf{L4} & \textbf{L5} & \textbf{L6} & \textbf{L7} & \textbf{L8} \\
\midrule
r$\cap$box     & $0.888 \pm 0.017$ & $0.737 \pm 0.098$ & $0.648 \pm 0.069$ & $0.648 \pm 0.021$ & $0.623 \pm 0.017$ & $0.559 \pm 0.054$ \\
c$\cap$box     & $0.650 \pm 0.064$ & $0.582 \pm 0.093$ & $0.707 \pm 0.058$ & $0.657 \pm 0.017$ & $0.636 \pm 0.017$ & $0.587 \pm 0.053$ \\
r,$\lnot$box   & $0.390 \pm 0.048$ & $0.443 \pm 0.076$ & $0.325 \pm 0.035$ & $0.339 \pm 0.017$ & $0.329 \pm 0.016$ & $0.351 \pm 0.066$ \\
c,$\lnot$box   & $0.306 \pm 0.062$ & $0.342 \pm 0.063$ & $0.383 \pm 0.068$ & $0.326 \pm 0.016$ & $0.319 \pm 0.018$ & $0.333 \pm 0.062$ \\
box            & $0.560 \pm 0.070$ & $0.332 \pm 0.093$ & $0.379 \pm 0.047$ & $0.340 \pm 0.022$ & $0.298 \pm 0.022$ & $0.334 \pm 0.074$ \\
stack          & $0.238 \pm 0.066$ & $0.105 \pm 0.035$ & $0.069 \pm 0.014$ & $0.053 \pm 0.014$ & $0.036 \pm 0.014$ & $0.224 \pm 0.061$ \\
band           & $0.097 \pm 0.069$ & $0.049 \pm 0.030$ & $0.071 \pm 0.013$ & $0.057 \pm 0.013$ & $0.040 \pm 0.016$ & $0.223 \pm 0.075$ \\
none           & $-0.062 \pm 0.055$ & $0.014 \pm 0.014$ & $0.009 \pm 0.007$ & $-0.016 \pm 0.009$ & $-0.022 \pm 0.010$ & $0.117 \pm 0.067$ \\
\bottomrule
\end{tabularx}
\caption{Mean cosine similarity with std across cell probe vectors trained to predict valid candidates in a cell. 
Scores are grouped by probes sharing a digit and a certain type of substructure, e.g. r$\cap$box means that two probes share a digit, a row and a box.}
\label{tab:probe-cosine-similarities-stats}
\end{table}

\subsection{MSE and AUC Scores for Transfer Across Timesteps}
\label{app:transfer-steps}
\begin{center}     
     \begin{minipage}{0.48\textwidth}
         \centering
         \includegraphics[width=\textwidth]{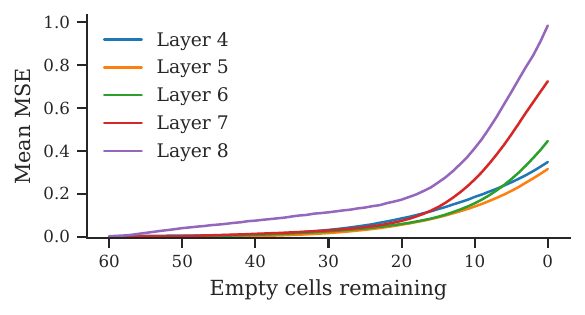}
     \end{minipage}
     \hfill
     \begin{minipage}{0.48\textwidth}
         \centering
         \includegraphics[width=\textwidth]{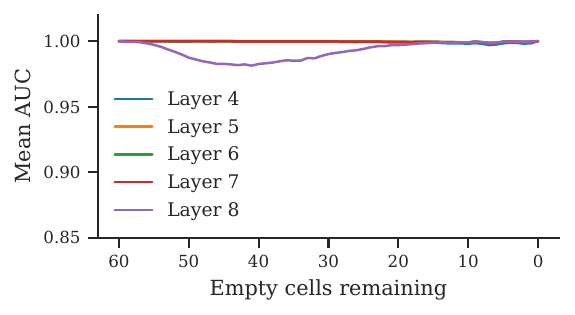}
     \end{minipage}
     
     \captionof{figure}{
Cross-position transfer diagnostics complementing \cref{fig:transfer} (right).
Frozen substructure-candidate probes trained at the \cluesend token are evaluated 
at every subsequent position of the solving trace, plotted against the number of 
empty cells remaining at that position. Curves are means over 1{,}280 held-out 
puzzles, with one curve per probe layer (L4--L8).
\textbf{(Left)} Mean squared error rises sharply as the puzzle is solved, 
particularly for L8 which approaches $\approx\!1.0$ at the end of 
the trace.
\textbf{(Right)} Mean ROC-AUC of the same probes stays at $\approx\!1.0$ for 
L4--L7 across the entire trace, with only L8 dipping marginally (to $\approx\!0.98$) 
in the mid-trace region.
The contrast between the two panels confirms that
the apparent degradation in exact-match accuracy across the trace is a calibration artefact of the probes' fixed 
biases, not a loss of the underlying substructure directions, which remain 
linearly separable throughout generation.
}
     \label{fig:probes-mse}
\end{center}

\subsection{Causal Ablation Using Substructure Probe Vectors}
\label{app:structure-present-patch}

\begin{table}[h]
\centering
\begin{tabularx}{\textwidth}{lrrrr}
\toprule
Layer & Logit drop & Patched logit & Valid top-1 & Changed top-1 \\
\midrule
L0 & $13.978 \pm 0.311$ &  2.845 & 0.723 & 0.987 \\
L1 & $ 8.724 \pm 0.293$ &  8.099 & 0.690 & 0.953 \\
L2 & $15.054 \pm 0.295$ &  1.769 & 0.723 & 0.993 \\
L3 & $23.727 \pm 0.332$ & -6.904 & 0.763 & 0.993 \\
L4 & $24.766 \pm 0.313$ & -7.943 & 0.727 & 0.997 \\
L5 & $22.124 \pm 0.285$ & -5.301 & 0.650 & 0.997 \\
L6 & $21.775 \pm 0.264$ & -4.952 & 0.597 & 0.997 \\
L7 & $19.235 \pm 0.243$ & -2.413 & 0.607 & 0.993 \\
\bottomrule
\end{tabularx}
\caption{
Per-layer results of the counterfactual patching intervention using substructure probes.
\emph{Logit drop}: mean drop in the logit of the target token relative to the clean run 
(mean clean logit 16.82), reported as mean $\pm$ standard error.
\emph{Patched logit}: mean absolute logit of the target token after patching.
\emph{Valid top-1}: fraction of patched runs whose new top-1 prediction is itself 
a valid placement on the original board.
\emph{Changed top-1}: fraction of patched runs whose top-1 differs from the clean 
top-1.}
\label{tab:structure-present-patch}
\end{table}
\newpage

\subsection{Substructure Organization of the Unembedding Vectors}

We note that the placement-token unembeddings are substructure-organized. 
Pairwise cosine similarities among the 729 unembedding vectors cluster by which features the tokens share — 
row, column, box, or digit — 
with near-identical similarity values within each cluster. 
For instance, all token pairs sharing a row and a digit have similarity $0.2$, and those sharing a row, 
a box and a digit $0.37$, while logically unrelated pairs are near-orthogonal. 

\begin{center}
  \includegraphics[width=\linewidth]{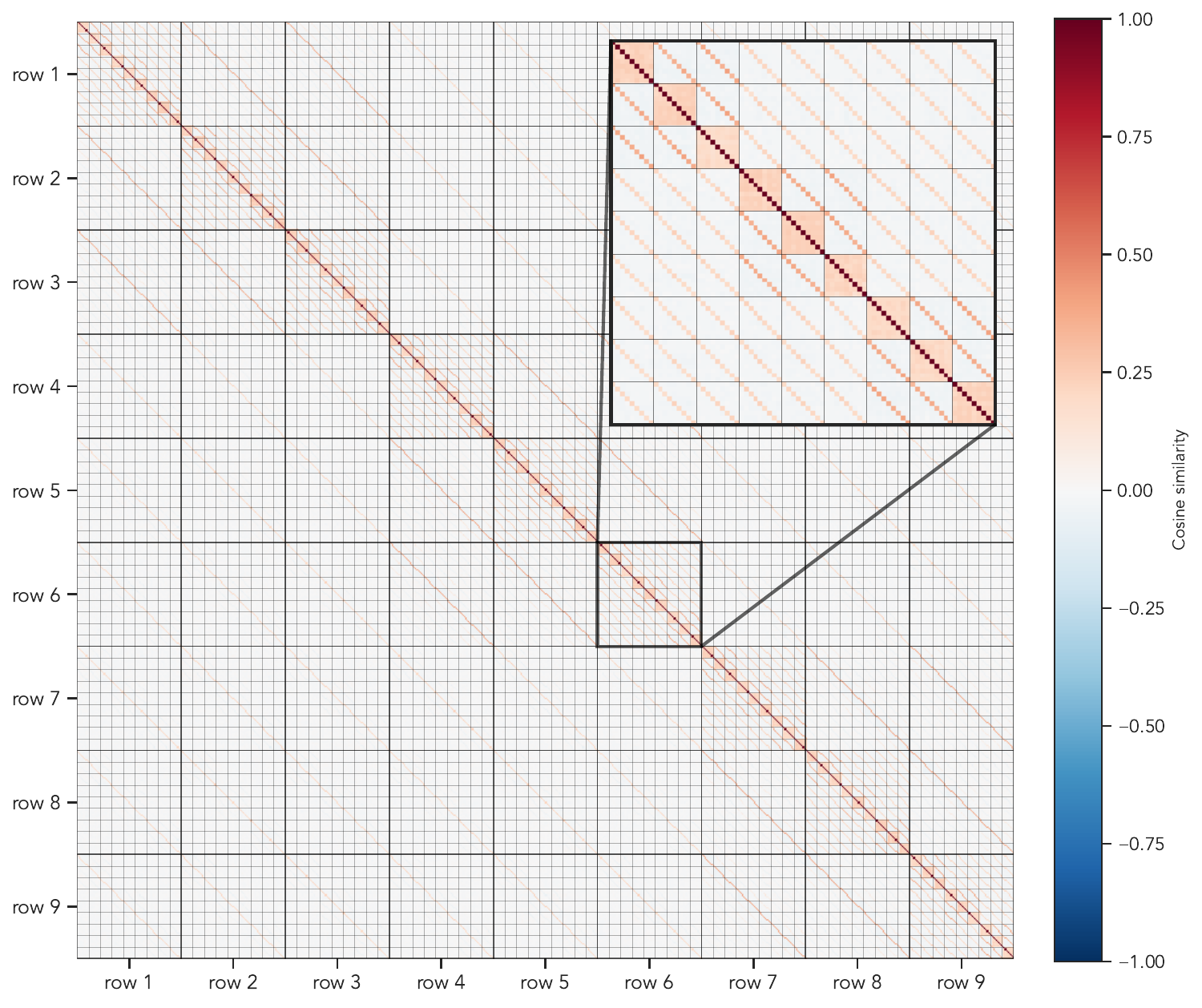}
  \captionof{figure}{Pairwise cosine similarity of the unembedding vectors.
  Each of the nine big cells is a row, each small square is a column, and the smallest squares are digits.
  Zoomed-in part shows cosine similarities of the unembedding vectors of row 6.}
\end{center}

\newpage

\section{Mid-Layer Attention}
\subsection{Mean Attetion From \cluesend Across All Heads}
\label{add:full-attention-patterns}

\begin{center}
  \includegraphics[width=\linewidth]{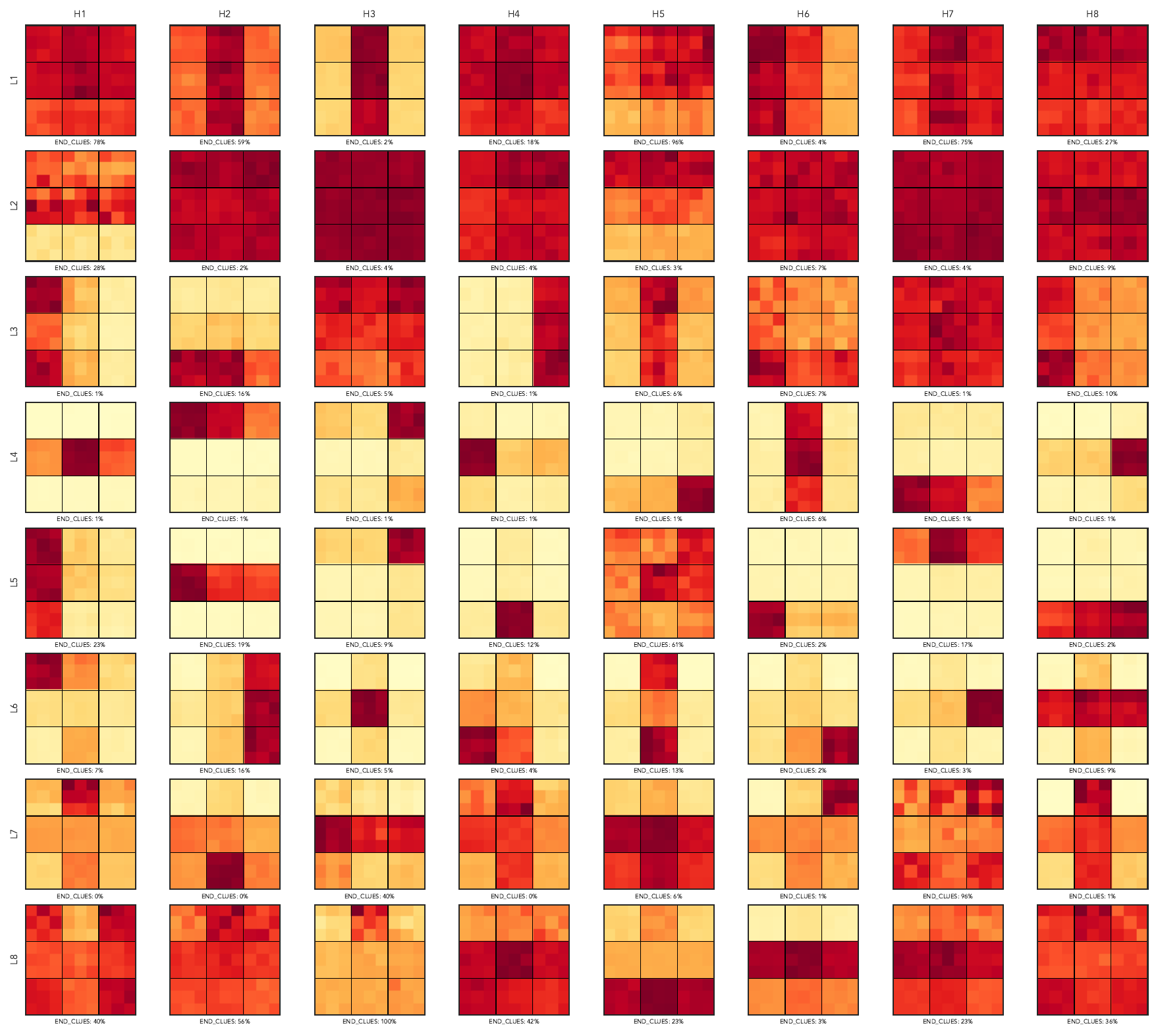}
  \captionof{figure}{Mean attention scores from \cluesend token to other tokens in the grid, averaged over all digits, computed over 6400 puzzles.}
\end{center}

\newpage

\subsection{Additional Examples of DLA for Substructure Aligned Heads}
\label{app:more-heads-dla}

\begin{center} 
     
     \begin{minipage}{0.8\textwidth}
         \textbf{(a)} \includegraphics[width=0.95\textwidth, valign=m]{fig_head_circuit_L4H6_cols4.pdf}
     \end{minipage}
     \\ \vspace{0.5em} 
     
     \begin{minipage}[b]{0.8\textwidth}
         \textbf{(b)} \includegraphics[width=0.95\textwidth, valign=m]{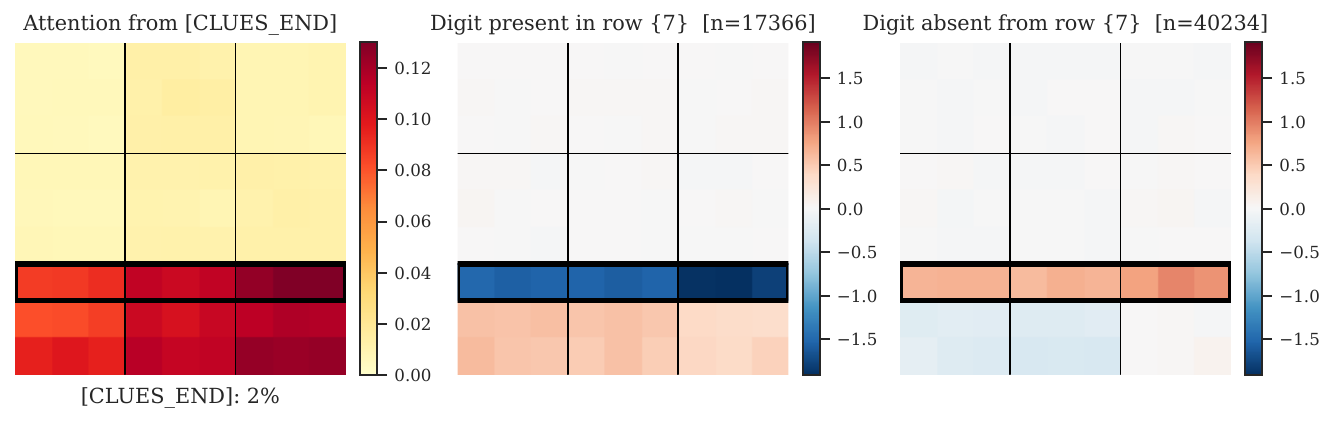}
     \end{minipage}
     \\ \vspace{0.5em}
     
     \begin{minipage}[b]{0.8\textwidth}
         \textbf{(c)} \includegraphics[width=0.95\textwidth, valign=m]{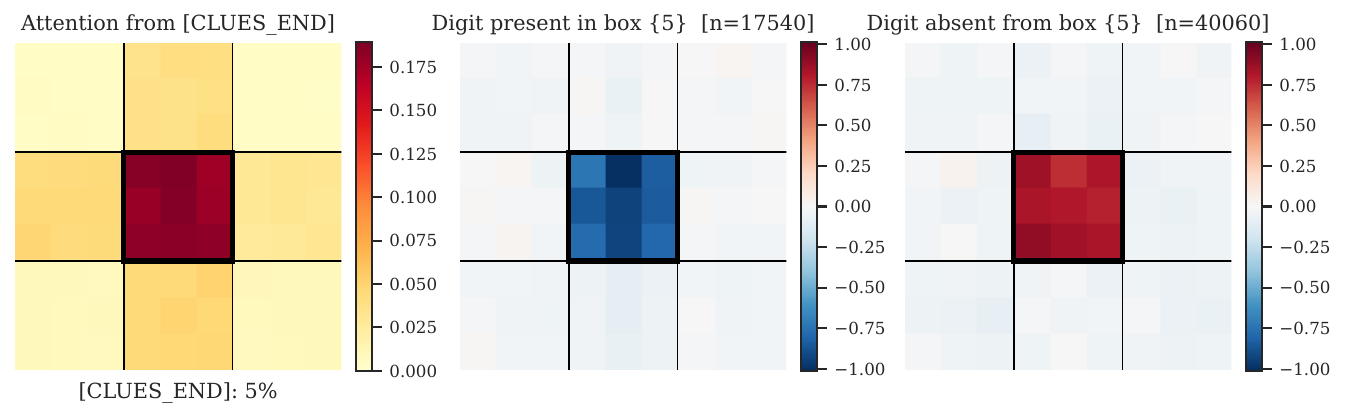}
     \end{minipage}
     
     \captionof{figure}{Substructure constraint elimination in mid-layer attention heads. 
     Attention maps (left) shows that heads are specialized for routing information 
     from specific substructures: (a) columns, (b) rows, and (c) boxes. 
     Direct Logit Attribution (middle and right) shows that each head 
     suppresses candidate logits for digits already present in its target substructure,
    while promoting absent digits.
    }
     \label{fig:more-heads-dla}
\end{center}

\section{NS Ciruit Materials}
\label{app:ns-dla}

\begin{table}[htpb]
\centering
\begin{tabular}{lrrrrr}
\toprule
\textbf{Statistic} & \textbf{Target Mean} & \textbf{Target Std} & \textbf{Other Mean} & \textbf{Other Std} \\
\midrule
\textbf{Mean} & 3.00 & 0.11 & $-0.03$ & 0.03 \\
\textbf{Std} & 0.72 & 0.04 & 0.01 & 0.01  \\
\textbf{Min} & 0.87 & 0.03 & $-0.05$ & 0.02  \\
\textbf{25\%} & 2.53 & 0.08 & $-0.03$ & 0.03 \\
\textbf{50\%} & 3.12 & 0.10 & $-0.03$ & 0.03 \\
\textbf{75\%} & 3.48 & 0.13 & $-0.02$ & 0.03 \\
\textbf{Max} & 4.48 & 0.23 & $0.00$ & 0.05\\
\bottomrule
\end{tabular}
\caption{DLA metrics for the $N=91$ identified naked single neurons. 
\textit{Target} metrics represent logit contributions to the 9 placement tokens of the neuron's associated cell. 
\textit{Other} metrics represent the global projection onto all 720 placement tokens of the remaining 80 empty cells.}
\label{tab:dla_metrics}
\end{table}



\end{document}